\pdfoutput=1

\documentclass[11pt]{article}

\usepackage[]{EMNLP2023}

\usepackage{times}
\usepackage{latexsym}
\usepackage{enumitem}
\usepackage{booktabs}
\usepackage{multirow}
\usepackage{tabularx}
\usepackage{graphicx}
\usepackage{verbatim}
\usepackage{amsmath}
\usepackage{xcolor}

\definecolor{cusviolet}{HTML}{E5CCFF}
\definecolor{cusindigo}{HTML}{CCE5FF}
\definecolor{cusblue}{HTML}{CCFFFF}
\definecolor{cusgreen}{HTML}{CCFFCC}
\definecolor{cusyellow}{HTML}{FFFFCC}
\definecolor{cusorange}{HTML}{FFE6CC}
\definecolor{cusred}{HTML}{FFCCCC}
\definecolor{cusskin}{HTML}{E6D0DE}

\usepackage{hyperref}
\hypersetup{
    }
    
\usepackage[T1]{fontenc}

\usepackage[utf8]{inputenc}
\usepackage{subfig}

\usepackage{microtype}

\usepackage{inconsolata}
\newcommand\tab[1][1cm]{\hspace*{#1}}

\title{ActionCOMET: A Zero-shot Approach to Learn \\ Image-specific  Commonsense Concepts about Actions }

\author{Shailaja Keyur Sampat \and Yezhou Yang \and Chitta Baral 
\\ \texttt{\{ssampa17, mpatel57, yz.yang, chitta\}@asu.edu} \\ 
         Arizona State University}

\begin{document}
\maketitle



\begin{abstract}

Humans observe various actions being performed by other humans (physically or in videos/images) and can draw a wide range of inferences about it beyond what they can visually perceive. Such inferences include determining the aspects of the world that make action execution possible (e.g. liquid objects can undergo pouring), predicting how the world will change as a result of the action (e.g. potatoes being golden and crispy after frying), high-level goals associated with the action (e.g. beat the eggs to make an omelet) and reasoning about actions that possibly precede or follow the current action (e.g. crack eggs before whisking or draining pasta after boiling). Similar reasoning ability is highly desirable in autonomous systems that would assist us in performing everyday tasks. To that end, we propose a novel multi-modal task to learn aforementioned concepts about actions being performed in images. We develop a dataset consisting of 8.5k images and 59.3k inferences about actions grounded in those images, collected from an annotated cooking-video dataset. We propose ActionCOMET\footnote{Code is available at \url{https://github.com/shailaja183/ActionConceptLearning.git}}, a zero-shot framework to discern knowledge present in language models specific to the provided visual input. We present baseline results of ActionCOMET over the collected dataset and compare them with the performance of the best existing VQA approaches.  

\end{abstract}

\section{Introduction}

From a very early age, humans are able to form a mental representation of surrounding objects through a combination of active learning (first-hand experience and imitation) and passive learning (reading from books, watching videos, conversing with other people, etc.). As we grow older, we are able to make more complex and abstract inferences, including the ability to predict intentions, anticipate future states of the world, and respond to imaginary situations \cite{moore2013development}. Over time and with experience, we build a cumulative knowledge repository of reusable concepts that enables our interaction with the world in a seamless manner. 

Moreover, our interaction with the world is heavily goal-driven. Humans capitalize on their knowledge about objects (such as their properties, relationship with other objects, and affordances) and perform necessary actions over them to accomplish desired goals. That being said, a significant amount of commonsense knowledge we use in our day-to-day life revolves around actions. For example, we can easily perceive when a stack of dishes or a Jenga tower will topple; we can predict whether the object is firmly attached to the surface or free to be lifted; and we can estimate to what extent the grocery bag should be filled so that it does not tear or crush the contents \cite{battaglia2013simulation}.

Commonsense reasoning about actions is crucial for humans to navigate in the world, as highlighted in several works- (i) to visualize what sequence of actions will lead us to the goal and to reduce exploration \cite{murugesan2021text}; (ii) to perform look-ahead planning and determine how current actions will affect future states of the world \cite{JMLR:v17:13-584}; (iii) to explain observations in terms of what actions may have taken place \cite{baral2010reasoning}; and (iv) to diagnose faults i.e. identifying actions that may have caused undesirable situations \cite{baral2010reasoning}. 

As we are developing autonomous agents that can assist us in performing everyday tasks, they would also have to interact with complex environments and possess the commonsense to resolve situations that may arise. The importance of this research problem has been identified since the early days of AI \cite{mccarthy1960programs, mccarthy1963situations, McCarthy1969-MCCSPP}. Since then, there has been a lot of progress and active efforts in curating large-scale commonsense resources and in building systems that can tackle varied aspects of action reasoning (we discuss them in detail under Section \ref{sec:lit}). However, state-of-the-art models (including large language models) face numerous challenges with tasks that require reasoning, which is quite trivial for humans \cite{li-etal-2022-systematic, valmeekam2022large, bian2023chatgpt}. One key reason for this performance gap between humans and AI models is the coarse-grained nature of most existing commonsense benchmarks e.g., a person cooks if they are hungry, a person calls the police in order to be lawful, a person will smile if someone complements them \cite{sap2019atomic}. 

In our opinion, commonsense concepts related to actions are much more complex behind-the-scenes than surface-level correlations, which in turn require consolidated physical, spatial, temporal, visual, and causal understanding. 
Such concepts include \textit{pre-conditions} (e.g. necessary ingredients and kitchen equipment), \textit{executability} (e.g. liquid objects can be poured, solid objects can be cut), \textit{effects} (e.g. straining will remove excess water), \textit{high-level goals} (e.g. to make an omelet) and \textit{temporal dependency} (e.g. crack eggs before beating or whisking). To address this challenge, we make the following contributions in this paper:

\begin{itemize}[leftmargin=*,noitemsep,topsep=0pt]
\item We propose a novel multi-modal task to learn concepts about actions (pre-conditions, effects, goals, and past/future actions) from images. 
\item We create a dataset of 8.5k images and 59.3k inferences about actions grounded in images, collected using an annotated cooking-video dataset.
\item  We propose ActionCOMET, a zero-shot approach to discern knowledge present in language models specific to the input images. 
\item We perform ablations using diverse input prompts, report quantitative/qualitative results of our approach and compare with the performance of state-of-the-art VQA systems. 

\end{itemize}

\section{Related Work}
\label{sec:lit}
In this section, we situate and compare our work with existing efforts spanning three research areas within Vision and Language (V\&L)- reasoning about actions, commonsense reasoning, and commonsense knowledge graphs. 

\subsection{Reasoning about Actions in V\&L}

From the V\&L research point of view, we identify four major kinds of tasks that involve reasoning about actions\footnote{We forbear action recognition task here, as it does not involve `reasoning' beyond pixel-level information.}- (i) \textit{Temporal Prediction} \cite{yiclevrer, sampat2021clevr_hyp, Patel2022CRIPPVQACR}, where provided a state of the world and an action which has taken place/to be performed, the task is to predict respective past/future state; (ii) \textit{Temporal Explanation} \cite{gao2018action, yeo2018visual, dalvi2019everything, park2019viewpoint, bisk2020piqa}, where given premise, the model selects/generates a plausible explanation that justifies causal nature of the premise; (iii) \textit{Goal-driven Planning} \cite{Bisk2018LearningIS, mattersim, Gokhale2019CVPRWorkshops, shridhar2020alfred, girdhar2020cater, yang2021visual, hong2021transformation}, where a model selects action(s) to achieve the given goal; and (iv) Temporal Sequencing \cite{yagcioglu2018recipeqa, wang-etal-2019-youmakeup}, where provided a set of actions, the model has to determine their correct order of execution. 

In this paper, we aim to learn commonsense about actions (specifically preconditions, effects, and high-level goals). Hence, our work falls under the Temporal Prediction task defined above. The works of \citet{yiclevrer, sampat2021clevr_hyp, Patel2022CRIPPVQACR} employ multiple choice question answering (QA) to predict the effects of simple actions (e.g. add, remove, change) performed over objects in a synthetic image/video. In contrast, we leverage cooking videos which are more diverse in terms of objects and complexity of incorporated actions. Moreover, the aforementioned works assume actions to be always executable and guaranteed to produce visually observable effects, which might not be the case in real-world situations. Our framework is able to reason about preconditions of actions and can successfully identify scenarios where actions cannot be performed. 

\subsection{Commonsense Reasoning in V\&L}
Commonsense reasoning is perceived as an important capability of AI agents, yet poses numerous challenges to the best existing models, which humans find trivial. To support the development of such systems and test their commonsense abilities, many large-scale benchmarks have been developed. Among them, SWAG \cite{zellers2018swag}, CosmosQA \cite{huang2019cosmos}, CommonsenseQA \cite{abs-1810-12885}, SocialIQA \cite{Sap2019SocialIC}, WinoGrande \cite{sakaguchi2021winogrande}, ProtoQA \cite{boratko2020protoqa}, and MC-TACO \cite{zhou2019going} support a wide range of text-based commonsense inference. Whereas,  KBVQA \cite{wang2017explicit}, FVQA \cite{wang2017fvqa} and VCR \cite{zellers2019recognition} datasets require models to have both visual understanding as well as commonsense reasoning capabilities. 

Most of the above datasets leverage span matching or multiple choice QA for evaluation. Though we agree upon the convenience of classification methods from an evaluation viewpoint, we prefer to model commonsense reasoning as a generative task in this work. In our opinion, training a model with classification objective limits the model's generalization ability for unseen scenarios and understanding about plausible nature of commonsense \cite{davis2015commonsense}. There are a few works on generative commonsense \cite{sap2019atomic, bosselut2019comet, park2020visualcomet} however they are better fit to commonsense knowledge graphs subtopic, hence we expand on them in Section \ref{sec:pointer}. 

Although all the above datasets tackle commonsense reasoning in some or another way, they differ in terms of scope and coverage. Some datasets focus on a particular kind of commonsense (e.g. SocialIQA and MC-TACO focus on social and temporal commonsense respectively) whereas others (e.g. CommonsenseQA, CosmosQA, VCR) tests systems on more than one commonsense dimensions. In this work, our goal is to learn commonsense about actions, which is novel to the best of our knowledge. However, we acknowledge that it may somewhat overlap with physical, social, spatial, temporal, and causal commonsense. Though our scope is limited to the cooking domain here, it is fairly rich (from a commonsense perspective) yet complex enough to be an accurate reflection of human's commonsense inference abilities. 

\subsection{Commonsense Knowledge Graphs}
\label{sec:pointer}

There has been a handful of attempts to unify vast amount of common knowledge in the form of semi-structured knowledge graphs (KGs). Examples of such KGs include WordNet \cite{miller1995wordnet}, VerbNet \cite{schuler2005verbnet}, DBPedia \cite{auer2007dbpedia}, ConceptNET \cite{speer2017conceptnet}, and WebChild \cite{tandon2014acquiring}. Specifically, the nodes in the KGs represent concepts and they are connected with other concepts with edges which come from a pre-defined set of relationships, popularly referred to as $($subject, relation, object$)$ triples. There exists a range of methods \cite{lv2020graph, zhong2019improving, murugesan2021text,  zhan2022pathreasoner, luo2021just, wu2022multi, Ye_2023_CVPR} that query concepts and/or relations stored in these knowledge bases and integrate the retrieved information to tackle a broad range of tasks. The need for human-in-the-loop curation of commonsense KGs, the requirement of exact match of concepts while querying KGs, limited generalization for novel concepts, and disconnect from other commonsense reasoning types (such as visual and social commonsense) remain major bottlenecks concerning this research area. 

Besides, there are knowledge bases such as Event2Mind \cite{rashkin2018event2mind}, ATOMIC \cite{sap2019atomic}, and COMET \cite{bosselut2019comet} developed to address some of the above challenges with traditional text-based KGs. They focus on relations between events (described as a short phrase or a sentence) rather than entities and use text generation methods for modeling. They also tackle social aspects of commonsense to some extent (such as people's intents, reactions, and attributes). VisualCOMET \cite{park2020visualcomet} extends and augments COMET's \cite{bosselut2019comet} framework to obtain commonsense inference about events and intents grounded in images, which is closest to our work in this paper (as shown in Figure \ref{fig:vcvsac}). 

In VisualCOMET \cite{park2020visualcomet}, data compilation is based on movie clips, therefore images incorporated and commonsense inferences are highly people-centric. Contrary, we believe that the task of commonsense inference is equally important for visual content where humans are secondary subjects (e.g. recipe videos, product assembly, or do-it-yourself tutorials). Hence, commonsense around objects, their attributes, affordances, actions that can be performed over them, and effects that will be produced as a result will be of key interest. We aim to learn such inferences from cooking-related annotated video datasets. Finally, our dataset is designed with the compositionality and generalization (of objects-actions) point of view, which was not the case with VisualCOMET.

\section{Task and Dataset Creation}
\label{sec:td}
\subsection{Task}
\label{sec:task}

\begin{figure*}
\centering
\includegraphics[width=\textwidth]{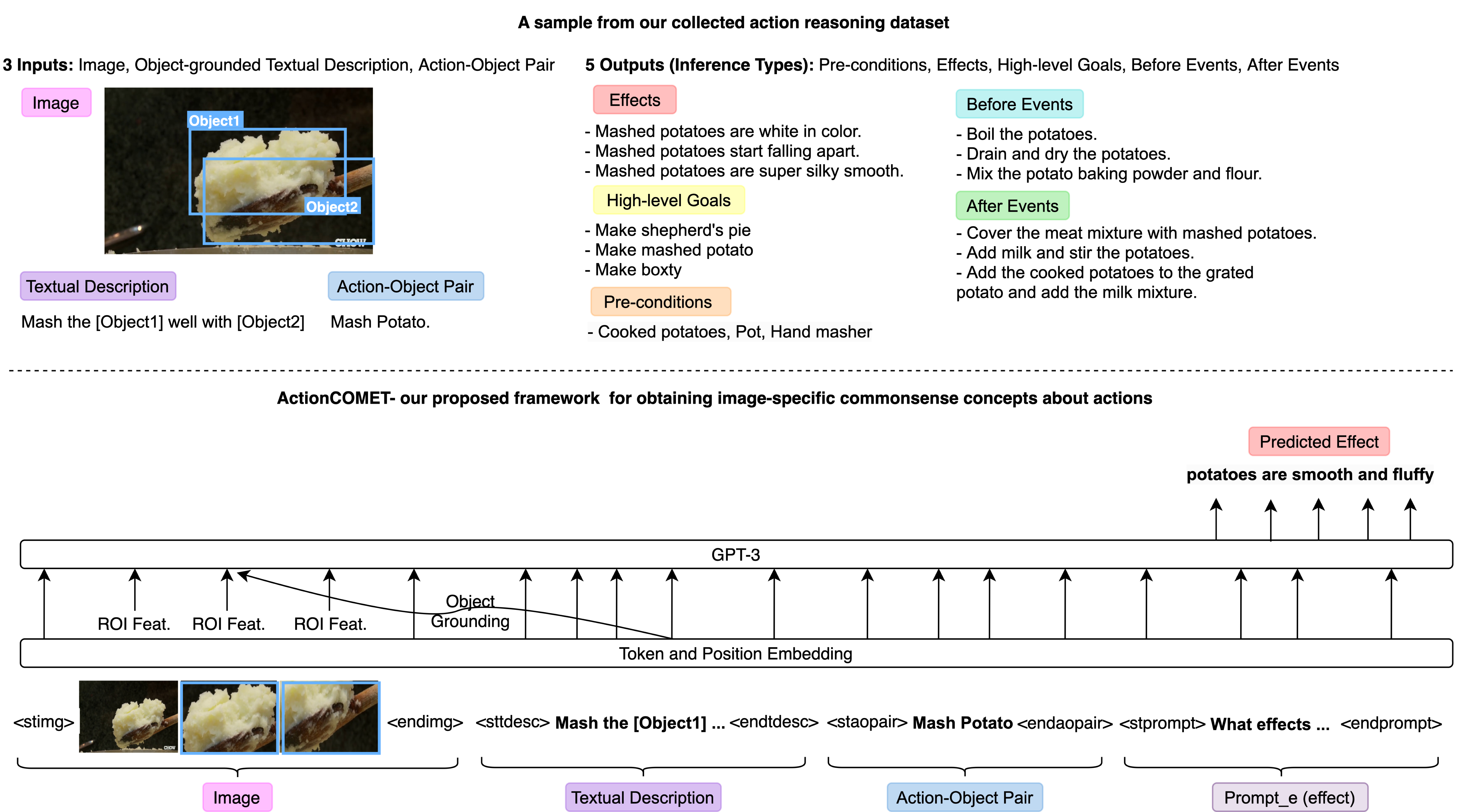}
\caption{Fine-grained action-centric commonsense generation addressed in this work: (top-left) Input to the model: an image, the textual description of an image and an action-object pair of interest (top-right) 5 types of action-related inference that we are interested in predicting using a vision-language model: Effects, High-level Goals, Pre-conditions, Before Events and After Events (bottom) The overview of the proposed ActionCOMET model demonstrating how different inputs are processed and the inference is generated}
\label{fig:vcvsac}
\end{figure*}
 
The task is to generate a visual commonsense to narrate the dynamic story of an \colorbox{cusskin}{image} from action-related aspects, as per the example shown in Figure \ref{fig:vcvsac}. The dynamic story consists of seven major commonsense inference components:
\begin{enumerate}
[leftmargin=*,itemsep=2pt,topsep=5pt]
    \item \colorbox{cusviolet}{Textual Description} This is equivalent to a dense image caption narrating key objects and actions being performed in the image. It is of form `cracking [Object1] using [Object2]' if egg and fork are detected in the image respectively. We refer to this as Object Grounding (OG) throughout this paper. It is an important first step as the generation of commonsense inference will be conditioned on these textual descriptions along with the visual features. 
    \item \colorbox{cusindigo}{Action-Object Pair} This component contains lemmatized primary verb and noun pair for the current event (e.g. crack egg). This can be quite similar to textual description but extracted from YouCook2 annotations instead of extracted from the input image. This component ignores secondary verbs and other supporting objects in the event which do not undergo the action directly (i.e. fork does not undergo cracking itself although it participates in the action of cracking eggs).  
    \item \colorbox{cusyellow}{High-level Goal} This component describes the motivations or intentions of a person/agent to perform a particular action over an object (e.g. action knead the dough is associated with the goal of baking the bread). The inclusion of this inference type is inspired by goal-driven human psychology to decide the course of action. 
    \item \colorbox{cusorange}{Pre-conditions} The set of actions you can perform on an object depends on the existence of other objects in the surrounding environment (e.g. presence of a knife or a scissor to cut) or because of physical properties of objects that receive the action (e.g. only liquid objects can undergo pouring). We intend to capture such common knowledge in order to equip models with the capability to determine if action execution is possible or not, specifically for unseen object-action pairs.
     \item \colorbox{cusred}{Effects} This inference describes the states of the object after an action is successfully performed over it. The state of the object can be described in the form of visual appearance (e.g. brown, puree, melted) or in terms of attributes it possesses (e.g. raw, soft, edible)  For example, peeling orange action removes the outer skin of an orange, or frying potato leads to the potato being golden or crisp. 
    \item \colorbox{cusblue}{Before Event}
    To reason about the dynamic situation of the image in the past, we use this inference type. This includes either the actions that might have taken place leading to the condition visible in the given image (e.g. peeling potato might have taken place before slicing potato if the skin is not visible in the image) or events involving other objects (e.g. water is boiled in a pot before putting pasta into it). 
    \item \colorbox{cusgreen}{After Event}  To reason about the dynamic situation of the image in the future, we use this inference type. This includes either the actions that are likely to take place after the condition visible in the given image for a specific goal (e.g. frying potato may take place after slicing if the goal was to make french fries), or events involving other objects (e.g. season the potato slices or coat them with corn flour). 
\end{enumerate}


\noindent The commonsense reasoning task is plausible in most cases. In other words, one can come up with multiple reasonable commonsense conclusions depending on the information provided, assumptions they make, and the past knowledge \citet{davis2015commonsense}. For example, the action cut a cauliflower can be associated with many possible high-level goals such as- to make a cauliflower steak, curry, pizza crust, and many more. Humans favor some reasoning/justication over other plausible alternatives depending on their personal preference or their knowledge of dishes that use cauliflower as an ingredient. Similar reasoning can be applied to other inference types. To account for plausibility, we frame each of the above commonsense components as a set of one or more inferences.

\subsection{Dataset Collection}
\label{sec:collect}
VisualCOMET \cite{park2020visualcomet} data was completely crowd-sourced by human workers, which turned out to be quite expensive\footnote{\$4/image paid to AMT workers * \$60k images = \$240k total cost of data collection reported in \cite{park2020visualcomet}} as well as would be a time-consuming process. Contrary, here, we leverage existing annotated video datasets as a resource to collect data for our task. Specifically, we set up a pipeline based on multiple off-the-shelf NLP tools (including co-reference resolution, dependency parser, and reading comprehension models) to collect a variety of commonsense knowledge about actions, with minimal human intervention. 

\begin{figure}[h!]
\includegraphics[width=\linewidth]{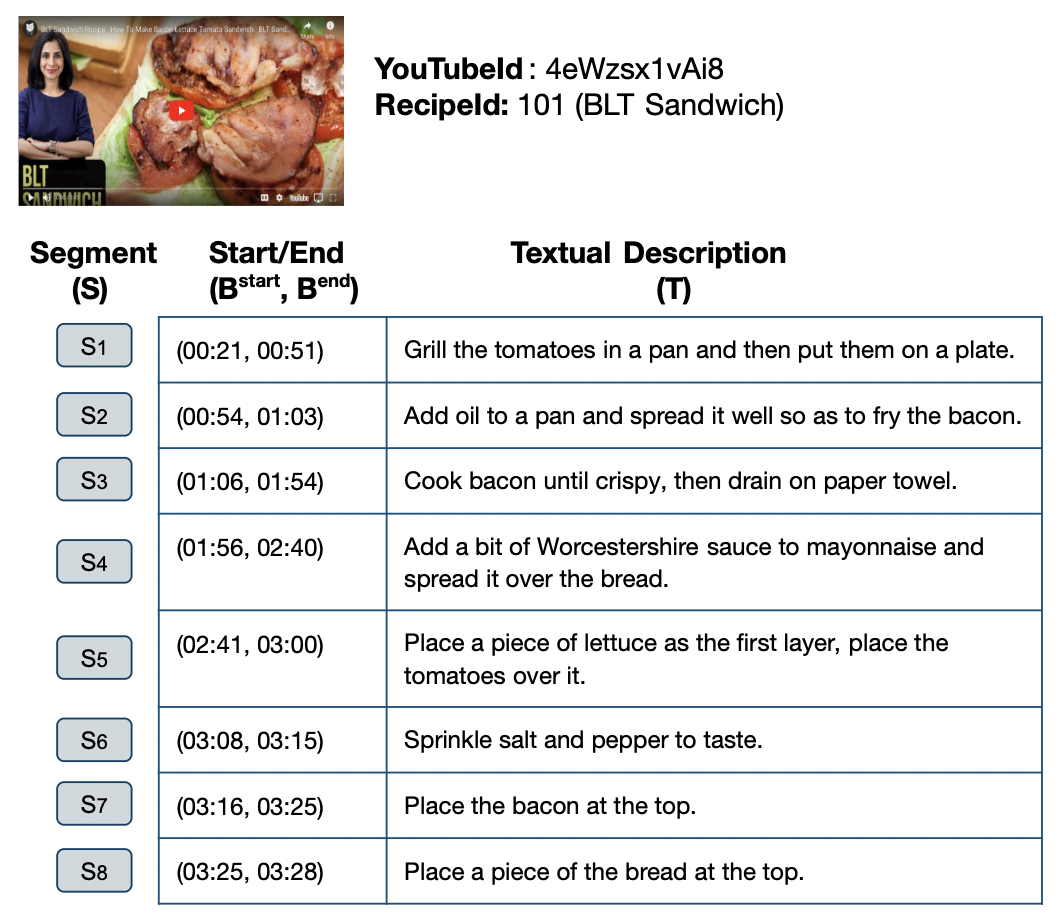}
\caption{Sample \href{https://www.youtube.com/watch?v=4eWzsx1vAi8}{4eWzsx1vAi8} from train+val partition of YouCook2 dataset with annotated procedure steps}
\label{fig:yc2}
\end{figure}

We use the existing YouCook2 dataset \cite{ZhXuCoAAAI18}, which is a collection of large task-oriented instructions (about the cooking domain), developed with an objective to improve video understanding models. We use their train+val partition, which contains annotations about 1790 videos available on YouTube covering 89 unique recipes. Each video in the dataset is divided into multiple procedure steps or segments $\{$S$_1$, S$_2$,..., S$_N$$\}$, which represent key events occurring in the video. Each segment S$_i$, i $\in$ $\{$1,2,..,N$\}$ is annotated with respective temporal boundaries (i.e. B$_i^{start}$ and B$_i^{end}$) and has a descriptive sentence T$_i$ associated to it, as per example shown in Figure \ref{fig:yc2}. They also provide bounding box annotations for each segment O$_i$ = $\{$o$_i^1$,  o$_i^2$, ..., o$_i^M$$\}$ if there were M different objects (o) appearing in the segment S$_i$ \cite{ZhLoCoBMVC18}. There is a RecipeID assigned to each video, denoting the type of the recipe (among 89). 


\begin{figure*}
\includegraphics[width=\linewidth]{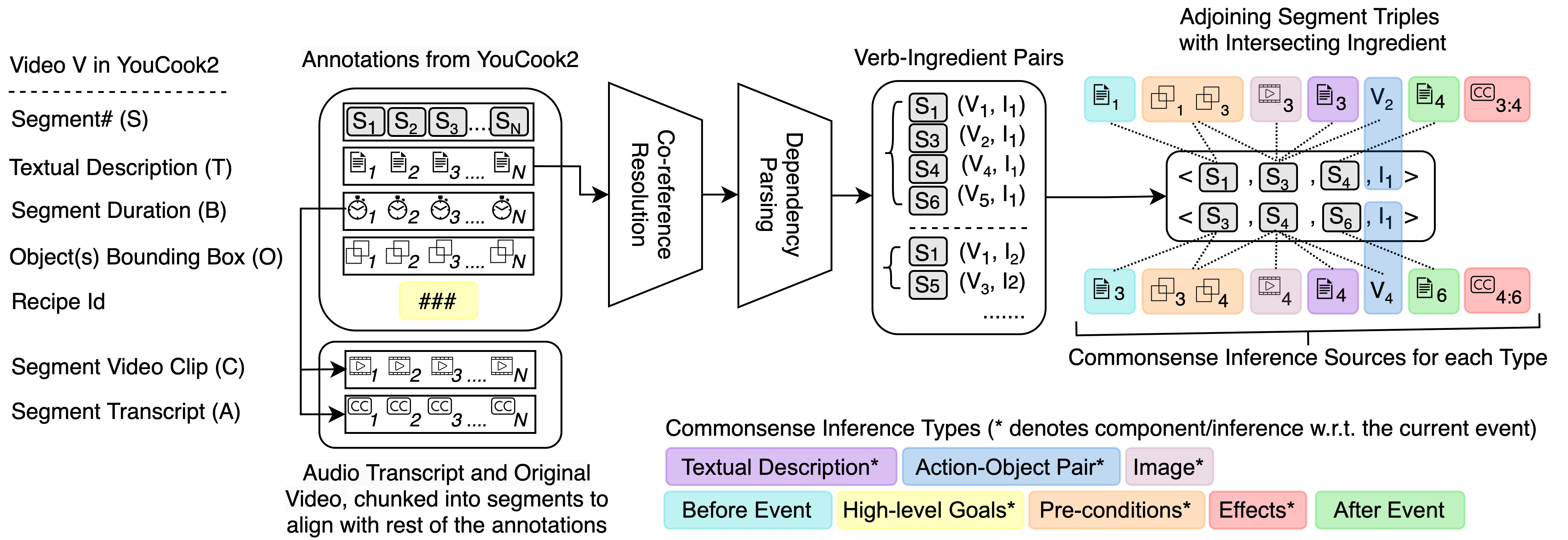}
\caption{Our data preparation pipeline that leverages YouCook2 \cite{ZhXuCoAAAI18} annotations to extract commonsense inference about actions (best viewed in color)}
\label{fig:dp}
\end{figure*}

The YouCook2 dataset \cite{ZhXuCoAAAI18} does not provide actual videos in their dataset. So we first use yt-dlp\footnote{\scriptsize{\url{https://github.com/yt-dlp/yt-dlp}}} python package to obtain video-only data and youtube-transcript-api\footnote{\scriptsize{\url{https://pypi.org/project/youtube-transcript-api/}}} to obtain the corresponding audio transcript. Afterward, we segment the obtained video into clips $\{$C$_1$, C$_2$,..., C$_N$$\}$ and audio into $\{$A$_1$, A$_2$,..., A$_N$$\}$ using segment duration (B$^{start}$, B$^{end}$) so that we can use them in sync with rest of the YouCook2 annotations. During this step, we could not access 189 videos as they are no longer available on YouTube and 395 transcripts due to video unavailable or closed captioning turned off by the user. For the remaining 1601 videos, we process them through the pipeline shown in Figure \ref{fig:dp}. 

As a first step, we resolve co-references present in textual descriptions T$_i$. For the BLT Sandwich example (Figure \ref{fig:yc2}), segments S$_1$, S$_2$ and S$_4$ and S$_5$ contain co-references (them/it) which refer to tomatoes, oil, mayonnaise, and lettuce respectively. In particular, we use F-coref's \citet{Otmazgin2022FcorefFA}  Spacy component. We further pass textual descriptions (T$_i$ with co-references resolved) through the dependency parser  to extract all  (Verb, Ingredient) pairs that directly connect in a dependency tree. In the given example, from S$_1$ we can obtain two such pairs- (Grill, Tomato) and (Put, Tomato). This process is repeated for all the segments in a video. However, in this step, we only keep pairs where both verb and ingredient have at least 10 occurrences (independent of each other) across the train+val partition. We propose to do such filtering to reliably collect commonsense inferences around verbs/ingredients and avoid any noise or rare utterances. We use Spacy's transformer-based en\_core\_web\_trf pipeline \cite{honnibal2020spacy}, which is state-of-the-art for dependency parsing. 

After obtaining all verb-ingredient pairs based on the above criteria, we group them by ingredients and sort them in ascending order by segment number to make sure their temporal ordering is preserved. For each ingredient, we collect adjoining segment triplets $<$S$_i$, S$_j$, S$_k$$>$ such that i$<$j$<$k. Here, S$_j$ would be considered as a current event, S$_i$ and S$_k$ being past and future events respectively. For the BLT sandwich example, the ingredient `bacon' appears in segments S$_1$, S$_2$, and S$_7$ hence we can create a segment triplet $<$S$_1$, S$_2$, S$_7$$>$. This denotes there are three actions fry $\rightarrow$ cook $\rightarrow$ place is performed over the bacon. If we consider `cook bacon' (S$_2$) as a current event, `fry bacon' (S$_1$) and `place bacon' (S$_7$) would be the past and future events respectively, for the `bacon' object. 

It is also possible that the ingredient may undergo processing in more than three segments. For example, consider the situation shown in Figure \ref{fig:dp}, where there are four segments S$_1$, S$_3$, S$_4$, S$_5$ which has the presence of ingredient I$_1$. Here, we can obtain two adjoining segment triplets i.e. $<$S$_1$, S$_3$, S$_4$$>$ and $<$S$_3$, S$_4$, S$_6$$>$. More generally, if an ingredient occurs in K different segments, where K$<$N, then we would  obtain K-2 segment triplets. For each triplet obtained, we use other annotations $\{$T, O, C, A, RecipeID$\}$ (either from YouCook2 or crawled information) to collect an image corresponding to the current event and commonsense inferences described in Section \ref{sec:task} as follows;

\begin{enumerate}
[leftmargin=*,noitemsep,topsep=0pt]
    \item \colorbox{cusviolet}{Textual Description} T$_j$ with object grounding corresponding to segment S$_j$ would be considered as ground-truth textual description of the current event. For testing instances, we will leverage image captioning systems to obtain textual description. 
    \item \colorbox{cusindigo}{Action-Object Pair} Verb V and Ingredient I associated with segment S$_j$ would be considered as an action-object pair for the current event. 
    \item \colorbox{cusyellow}{High-level Goal} We find the recipe name corresponding to the RecipeID (as per mapping shown in the Table \ref{tab:ridname}) and use the text template `Make/Cook/Prepare <recipe name>' for this inference type. 
    \item \colorbox{cusskin}{Image} We capture the middle frame of the video clip C$_j$ corresponding to segment S$_j$, which is a visual representation of the action-object pair for the current event.    
    \item \colorbox{cusorange}{Pre-conditions} We use bounding box annotations O$_i$ and O$_j$  corresponding to past event and current event respectively to form pre-condition inference. For example, consider the past event `potato is peeled using a peeler' and the current event `potato is cut into pieces using a knife over a chopping board'. In this case, $\{$potato, peeler, knife, chopping board$\}$ form pre-conditions for the current event. 
    \item \colorbox{cusred}{Effects} We rely on audio transcripts between current event and future event (A$_{i:j}$) for this purpose. Typically, in cooking videos, the person demonstrating a recipe acts first (i.e. current event) and then (but before the next event) describes how should the object look  after the action, in terms of color, shape, or other attributes. Therefore, we feed this audio transcript to a reading comprehension model and ask templated questions such as `What color/texture/shape is $<$ingredient$>$?' and `What attribute/property is related to $<$ingredient$>$?'. We use RoBERTa-base model \cite{liu2019roberta}, fine-tuned over SQuAD2.0 (including unanswerable questions) \cite{rajpurkar2018know} to collect effect annotations.
    \item \colorbox{cusblue}{Before Event} T$_i$ corresponding to segment S$_i$ (pre-cursor to the current event S$_j$) is used as a textual description of the before event.
    \item \colorbox{cusgreen}{After Event} T$_k$ corresponding to segment S$_k$ (following event to the current event S$_j$) is used as a textual description of the after event.
\end{enumerate}

As a final step, we combine inference from multiple videos which have common action-object pair to support the plausible nature of commonsense.  
For example, `mash potato' was a part of three different recipes (shepherd's pie, mashed potato, and boxty), hence we combine their inference across each type as shown in Figure \ref{fig:vcvsac} (top-right portion).  

\subsection{Data Statistics/Partitions}

Table \ref{tab:stat} summarizes statistics about the data we obtained using the process outlined in Section \ref{sec:collect}. We acknowledge that the size of the resulting dataset is smaller in comparison with VisualCOMET \cite{park2020visualcomet}, but our pipeline is easily reusable and scalable, even for domains beyond cooking. More importantly, our data collection process is fully automated and does not require human annotation. 

\begin{table}[h!]
\centering
\begin{tabular}{@{}ll@{}}
\toprule
\textbf{Annotated Input Type}                                                                                                                                                                                                           & \textbf{\#Samples} \\ \midrule
Videos & 1601 \\ 
Images$^*$                                                                                                                                                                                                                              &      8522     \\
Textual Descriptions$^*$ & 8522 \\
Recipe Types  & 89 \\

\begin{tabular}[c]{@{}l@{}}Actions-Objects$^*$ ($\geq$10 occurrences) \\ \tab Unique Objects\\ \tab Unique Actions 
\end{tabular}                                                                                                                                    &     \begin{tabular}[c]{@{}l@{}} \\ 176 \\ 93 
\end{tabular}                    \\ 
\begin{tabular}[c]{@{}l@{}}Commonsense Inferences\\ \tab High-level Goals$^*$\\ \tab Pre-conditions$^*$\\ \tab Effects$^*$\\ \tab Before Events\\ \tab After Events\end{tabular} &     \begin{tabular}[c]{@{}l@{}} \\ 10341 \\ 17209 \\ 6428 \\ 12665 \\ 12665
\end{tabular}               \\ \bottomrule
\end{tabular}
\caption{Statistics about ActionCOMET data ($^*$ denotes component/inference with respect to the current event)}
\label{tab:stat}
\end{table}


\begin{table*}
\centering
\begin{tabular}{@{}llllllll@{}}
\toprule
\textbf{Modalities related to visual inputs}                         &  \textbf{B} & \textbf{M} & \textbf{C} & \textbf{A@50} & \textbf{Unique} & \textbf{Novel} \\ \midrule
Image                                                &    7.34     &    8.12    &     6.55  &     13.60   &    23.14    &     45.30  \\
Image + OG                                           &   9.02     &    10.45    &    8.67   &     16.37   &      26.30  &   48.26    \\ 
AO Pair                                                                     &    6.77    &  9.01      &   7.66    &          11.35     &    32.36  &  56.12 \\
TextDesc                                                                        &   7.43        &  8.74      &  6.80     &  13.52      &     34.00   &   53.18    \\
AO Pair + TextDesc                                                                 &   7.49     &   8.68     &    6.91   &    13.47    &    35.23    &  52.84    \\ 
Image + TextDesc                                                                &    10.32    &    11.18    &   9.85    &    18.55    &     37.12   &   54.11    \\
Image + AO Pair                             &    15.45    &    17.10 &   16.85    &    21.28    &     35.33   &   50.45    \\

Image + TextDesc + AO Pair                                                            &   15.76     &     17.08   & 16.77       &     22.01   &    34.61    &    51.06   \\
Image + TextDesc + OG                                                &  \textbf{17.44}      &  \textbf{18.56  }    &    \textbf{17.20 }  &  \textbf{24.08 }     &   \textbf{36.31}     &     \textbf{49.56 } \\
Image + TextDesc + AO Pair + OG  &         \textbf{17.21}      &  \textbf{18.67 }    &    \textbf{17.35 }  &  \textbf{23.57 }     &   \textbf{37.31}     &     \textbf{51.56 } \\
\bottomrule
\end{tabular}
\caption{Ablations of baseline model from visual modalities perspective. Above results are for the Validation set using metrics BLEU-2 (B), METEOR (M), CIDER (C), Acc@50, Uniqueness and Novelty of generated inference. TextDesc is equivalent to the caption of the image, AO pair denotes ground-truth action-object pair shown in the image and OG denotes use of object grounding.}
\label{tab:abl1}
\end{table*}

\section{Model Description}
\label{sec:exp}

For each image in our dataset, we obtain sequence of visual embeddings v representing the image and objects detected in the image, grounded textual description t, action-object pair p, and inference type r. Then, we wish to generate a set of possible inferences H = $\{$s$^r_1$, s$^r_2$, ...s$^r_{|H|}$$\}$.

\noindent\textbf{Feature Extraction} $\frac{}{}$ \\
\noindent The sequence of visual representations V consists of a representation of the whole image and an additional representations for each object detected in the image. We use Region of Interest (RoI) Align features \cite{he2017mask} from Faster RCNN \cite{ren2015faster} as our visual embedding and pass it through a non-linear layer to obtain the final representation for an image or each detected object. The final sequence of representations V = $\{$v$_0$,v$_1$,..v$_m$$\}$ where m is the number of objects detected. We assign special tags identifying each object detected in the image (e.g. Object1, Object2 as shown in Figure \ref{fig:vcvsac}) in our dataset. To use these tags, we introduce new object tokens, e.g. [Object1] in the vocabulary and create additional word embedding for these tokens. Then, we sum the visual representation for an object with the word embedding of the token referencing the object in the text. This way, our model has visually grounded information about the image. We refer to this as Object Grounding (OG), which is inspired from Person Grounding done by \cite{park2020visualcomet}. \\ $\frac{}{}$ \\
\noindent\textbf{Text Representation} $\frac{}{}$ \\
\noindent Transformer models used for language tasks use special separator tokens to enable better understanding of the input structure. Since our task involves textual information of different kinds (events, pre-conditions, effects and goals), we include special tokens for our language representation as well. Specifically, we append special token indicating the start and end of image (e.g. s\_img, e\_img), event, action-object pair, and inference fields. To generate inference statements, we use one of the three inference types (before, after, goal, effect, precondition) as the start token, depending on the desired dimension.
\\ $\frac{}{}$ \\
\noindent\textbf{Inference Generation} $\frac{}{}$ \\
\noindent We use pre-trained GPT-3 \cite{brown2020language} for natural language generation, conditioned on v, t, p. 
For each inference s$^r_{h}$ $\in$ H, our objective is to maximize P(s$^r_{h}$|v,t,p,r). Suppose inference s$^r_{h}$ is a sequence of l tokens i.e. $\{$w$^r_{h1}$, w$^r_{h2}$, ...w$^r_{hl}$$\}$. Then, we minimize the negative log-likelihood loss over inference instances in the dataset as follows;

\begin{equation}
L=-\sum_{i=1}^{l}\operatorname{log}P\Bigl(w_{hi}^{r}|w_{h<i}^{r}, r, p, t, v\Bigl)
\label{fig:eq1}
\end{equation}

While our dataset provides events associated with each image, it is impractical to assume the availability of this information on new images. We experiment with a more general version of our model which does not take t (textual description) and p (action-object pair) as input. Nonetheless, we can supervise such models to generate t and p in the training phase. If we denote the current event with $\{$t$\}$ = $\{$w$^t_{1}$, w$^t_{2}$, ...w$^t_{n}$$\}$ and action-object pair as $\{$p$\}$ = $\{$w$^p_{1}$, w$^p_{2}$, ...w$^p_{n}$$\}$ as a sequence of tokens, we apply the seq2seq loss on t, p (which we refer to as TP loss) as per Equation \ref{fig:eq1}. 
\\ $\frac{}{}$ \\
\noindent\textbf{Model Parameters} $\frac{}{}$ \\
\noindent We use Adam optimizer \cite{kingma2014adam} with a learning rate of 5e-5 and batch size of 32. Visual features for image and object embeddings use ResNet101 \cite{he2016deep} backbone pretrained on ImageNet \cite{deng2009imagenet}. We set the maximum number of visual features to 15. We use pre-trained GPT3-base model \cite{brown2020language} as our model architecture with maximum total sequence length as 64. For decoding, we follow \citet{park2020visualcomet} and use nucleus sampling \cite{holtzman2019curious} with p = 0.9, to allow diverse text generation contrary to beam search. 

$\frac{}{}$ \\
\noindent\textbf{Evaluation Metric} $\frac{}{}$ \\
\noindent Here, we describe the automatic evaluation measuring the quality of inference sentences. We first report the automatic metrics used in image captioning such as BLEU-2 \cite{papineni2002bleu}, METEOR \cite{denkowski2014meteor}, and CIDER \cite{vedantam2015cider} across the 5 inferences. Inspired by the metric in visual dialog, we also use perplexity score to rank the ground truth inferences and inferences from the different image. We append negatives such that there are 50 candidates to choose from, rank each candidate using perplexity score, and get the average accuracy of retrieved ground truth inferences (A@50). Note that perplexity is not necessarily the perfect measure to rank the sentences, but good language models should still be able to filter out inferences that do not match the content in image and event at present. Lastly, we measure the diversity of sentences, so that we do not reward the model for being conservative and making the same predictions. We report the number of inference sentences that are unique within the generated sentences divided by the total number of sentences (Unique), and the number of generated sentences that are not in the training data divided by the total number of sentences (Novel). To capture the semantic diversity, we replace the predicted object tags with the same tag when calculating the above diversity scores.

\section{Experiments and Results}

\subsection{Ablation Study through Input Combinations} 
\noindent In our experiments, we fix the  model architecture shown as per Figure \ref{fig:vcvsac} and further described in Section \ref{sec:exp}. However, we do ablations on different combinations of the inputs available, which includes action-object pair (AO Pair), textual description of current event (TextDesc), and Image. We also measure the impact of object grounding (OG) for cases where Image modality is incorporated. As a result, we obtain 10 different combinations (as listed in Table \ref{tab:abl1}). The model is trained with the seq2seq objective mentioned in Equation \ref{fig:eq1} for each case, and we mask out the visual and/or textual input based on the ablation of interest. For each modality combination, we evaluate generated inferences on our collected data using BLEU-2 (B), METEOR (M), CIDER (C), Acc@50, Uniqueness and Novelty metrics.
\\ \\
\noindent\textbf{Quantitative Results:}
The results of the ablation study described above are summarized in Table \ref{tab:abl1}. We make the following observations: First, model that incorporates both visual and textual (Image + TextDesc + AO pair + OG) modalities outperform models trained with only one of the modalities (TextDesc + AO pair; Image + OG) in every metric, including retrieval accuracy and diversity scores. This indicates that the task requires  visual information is crucial and by incorporating textual descriptions relavant to the image can help to achieve higher quality inference. This reinforces the fact that the task at hand is of multi-modal nature.

Second, incorporation of action-object pair (AO pairs) seems to be more effective in comparison with textual description of the current event (TextDesc) when combined with images from the results perspective. As described earlier, AO pairs only incorporate one object at a time, which enables model to remain focused while generating inference about a given action-object combination. However, there might be another objects and actions that may passively participate in the action but do not undergo any effects themselves. Such broader context can only be understood through longer text modalities such as TextDesc. There is only a marginal performance improvement when both textual modalities are present together i.e. Image+AO Pair+TextDesc combination. Finally, when working with longer texts (TextDesc), the object grounding (OG) trick gives a performance boost to the model across all metrics.

\subsection{Ablation Study through Prompt Variations}
As shown in the bottom part of Figure \ref{fig:vcvsac}, we use a natural language prompt to instruct the model to specify which type of inference (among pre-condition, goal, effect, before action or after action) we want to obtain. In this section, we investigate the effect of using different prompts on the downstream performance. In particular, we consider four variations of each inference type- Assertive (sentence completion), Imperative, Interrogative, and Assertive (structured response). These prompt types were inspired by fundamental sentence constructions and language tones that are often used in human-languages i.e. English. Existing works \cite{yin2024should} have shown that LLMs are sensitive to such sentence variations and tones. The prompt variations incorporated in this study are summarized in Table \ref{tab:promptvar}. 

Assertive (sentence completion) prompt refers to instruction which provides model with some context about the type of desired inference (goal/pre-/effect etc.) and expects the model to complete the remaining sentence with appropriate inference. Pp1/Pe1/Pg1/Pb1/Pa1 prompts (referred in Table \ref{tab:promptvar}) are of assertive (sentence completion) kind. The imperative prompt refers to the sentence formulation which proposes model to return a particular response in an instructive or requesting tone. It typically starts with `Describe..' and covers prompts with suffix 2 (e.g. Pp2) in Table \ref{tab:promptvar}. The interrogative prompts, as the name suggests, are formulated as Wh-questions beginning with `What are some ...' instead of the sentence formulation. Finally, the assertive (structured response) prompt instructs the model to return itemized or enumerated answer instead of the free-form natural language response by explicitly stating the model to `List down' its response.

\begin{table*}
\centering
\resizebox{0.95\linewidth}{!}{%
\begin{tabular}{@{}ll@{}}
\toprule
\multicolumn{1}{c}{\textbf{Inference Type}}                           & \multicolumn{1}{c}{\textbf{Variations of prompts considered in this paper}}                                  \\ \midrule
\begin{tabular}[c]{@{}l@{}}Prompt\_p \\ (pre-conditions)\end{tabular} & \begin{tabular}[c]{@{}l@{}}Pp1: A set of concepts that are required to perform this action are\\ Pp2: Describe a list of necessary conditions required to execute this action\\ Pp3: What are some pre-requisites related to this action?\\ Pp4: List down things without which one cannot perform this action\end{tabular} \\ \midrule
\begin{tabular}[c]{@{}l@{}}Prompt\_e \\ (effects)\end{tabular}        & \begin{tabular}[c]{@{}l@{}}Pe1: Some results of performing this action include \\ Pe2: Describe what changes will be caused by performing this action\\ Pe3: What effects will be produced as a result of performing this action?\\ Pe4: List down the consequences if one performs this action\end{tabular}                \\ \midrule
\begin{tabular}[c]{@{}l@{}}Prompt\_g\\ (goal)\end{tabular}            & \begin{tabular}[c]{@{}l@{}}Pg1: Some objectives related to this action include\\ Pg2: Describe intents of people that are performing this action \\ Pg3: What are some high-level goals associated with this action?\\ Pg4: List down the recipes one can prepare which requires performing this action\end{tabular}        \\ \midrule
\begin{tabular}[c]{@{}l@{}}Prompt\_b\\ (before action)\end{tabular}   & \begin{tabular}[c]{@{}l@{}}Pb1: Some actions that person must have performed before this action are\\ Pb2: Describe which actions might have taken place in past\\ Pb3: What are some actions that typically take place before this action?\\ Pb4: List down some actions that preceded this action\end{tabular}            \\ \midrule
\begin{tabular}[c]{@{}l@{}}Prompt\_a \\ (after action)\end{tabular}   & \begin{tabular}[c]{@{}l@{}}Pa1: Some actions that person will perform after this action are\\ Pa2: Describe which actions are likely to take place in future \\ Pa3: What are some actions that typically take place after this action?\\ Pa4: List down some actions that will follow this action\end{tabular}             \\ \bottomrule
\end{tabular}}
\caption{Text prompt variations considered for GPT-3 for five inference types related to actions (pre-consitions, effects, goals, preceding and following actions) in our dataset}
\label{tab:promptvar}
\end{table*}

\begin{table*}
\centering
\resizebox{0.8\linewidth}{!}{%
\begin{tabular}{@{}llllllll@{}}
\toprule
\multicolumn{1}{c}{\textbf{Input components}} & \multicolumn{1}{c}{\textbf{\begin{tabular}[c]{@{}c@{}}Textual prompt\\ variation\end{tabular}}} & \multicolumn{1}{c}{\textbf{B}} & \textbf{M} & \textbf{C} & \textbf{A@50} & \textbf{Unique} & \textbf{Novel} \\ \midrule
\begin{tabular}[c]{@{}l@{}}Image + TextDesc \\ + AO Pair + OG\end{tabular} & Pp1 & 17.09 & 18.43 & 17.65 & 23.31 & 37.48 & 51.78 \\
 & \textbf{Pp2} & \textbf{18.33} & \textbf{20.41} & \textbf{19.19} & \textbf{24.28} & \textbf{36.78} & \textbf{50.55} \\
 & Pp3 & 18.19 & 18.67 & 18.48 & 23.01 & 39.22 & 51.99 \\
 & Pp4 & 17.57 & 19.42 & 19.79 & 24.04 & 37.44 & 50.35 \\ \cmidrule(l){2-8} 
 & Pe1 & 13.36 & 16.22 & 15.44 & 17.81 & 36.18 & 42.55 \\
 & Pe2 & 12.69 & 15.57 & 14.81 & 17.55 & 37.12 & 40.24 \\
 & Pe3 & 13.25 & 16.07 & 15.56 & 18.11 & 36.26 & 43.78 \\
 & \textbf{Pe4} & \textbf{15.34} & \textbf{17.34} & \textbf{16.66} & \textbf{18.27} & \textbf{35.42} & \textbf{43.35} \\ \cmidrule(l){2-8} 
 & Pg1 & 15.27 & 17.45 & 16.89 & 20.01 & 38.58 & 40.66 \\
 & Pg2 & 15.56 & 18.35 & 17.73 & 18.87 & 39.47 & 41.23 \\
 & Pg3 & 14.80 & 17.54 & 15.88 & 19.41 & 40.20 & 42.14 \\
 & \textbf{Pg4} & \textbf{16.23} & \textbf{18.89} & \textbf{17.77} & \textbf{20.24} & \textbf{37.20} & \textbf{43.67} \\ \cmidrule(l){2-8} 
 & Pb1 & 21.05 & 23.44 & 20.81 & 26.67 & 30.34 & 45.13 \\
 & Pb2 & 20.66 & 22.68 & 19.72 & 24.12 & 31.59 & 41.47 \\
 & \textbf{Pb3} & \textbf{21.34} & \textbf{24.05} & \textbf{21.67} & \textbf{25.78} & \textbf{30.56} & \textbf{43.25} \\
 & Pb4 & 19.08 & 21.82 & 22.01 & 25.12 & 31.22 & 42.42 \\ \cmidrule(l){2-8} 
 & Pa1 & 17.87 & 20.11 & 17.89 & 24.31 & 33.01 & 46.22 \\
 & Pa2 & 16.46 & 19.87 & 17.88 & 23.16 & 30.29 & 42.45 \\
 & \textbf{Pa3} & \textbf{19.03} & \textbf{22.41} & \textbf{20.21} & \textbf{25.37} & \textbf{34.11} & \textbf{45.27} \\
 & Pa4 & 17.56 & 21.61 & 19.33 & 24.03 & 31.29 & 43.65 \\ \bottomrule
\end{tabular}}
\caption{Ablations of baseline model from textual prompts perspective. Above results are for the Validation set using metrics BLEU-2 (B), METEOR (M), CIDER (C), Acc@50, Uniqueness and Novelty of generated inference.}
\label{tab:promptres}
\end{table*}

\noindent\textbf{Quantitative Results:}
The results of the ablation study regarding the prompt variations are summarized in Table \ref{tab:promptres}. Particularly, we start with the proposed model which considers both visual and textual modalities as input (i.e. Image + TextDesc with OG + AO pair), which was the best performing combination as reported in Table \ref{tab:abl1}. As evident from Table \ref{tab:promptres}, for different inference types, there are minor performance deviations when the prompt variations described above are used. For pre-condition, imperative prompt yields the best results. The model favors interrogative prompts to produce best inference for before and after events. The assertive (structured response) prompt is able to achieve the best performance for effect and goal inference. Based on the results, the use of assertive (sentence completion) prompts should be limited for all action-related inference dimensions.

Table \ref{tab:promptres} also provides insights related to how the model performs across each inference type. Across BLEU-2, METEOR, CIDER and Accuracy@50 metrics, the generated inference for pre-condition, before and after events is relatively better (with respect to the ground-truth) in comparison with effects and goals. Contrary to that, the uniqueness score for goals, effects and pre-conditions are higher. The novelty score for pre-conditions are significantly higher than the rest of the inference dimensions. However, low overall numbers throughout this table is indicating that state-of-the-art LLMs struggle to make concrete reasoning about actions, which humans can effortlessly perform in their day-to-day life.

\subsection{Human Evaluation}
We perform human evaluation with 3 individuals with respect to 100 samples (20 samples for each inference type) from the dataset we collected in this work. The evaluators were provided with the image, inference type of the interest and model generated natural language inference. They were advised to rate the sentence along two metrics:
\begin{itemize}[leftmargin=*,noitemsep,topsep=0pt]
    \item Correctness of the generated inference (choose from Incorrect / Partially-correct / Correct) i.e. provided the image and the desired inference type (e.g. effect), the participant indicates whether the generated inference was correct or not.
    \item Creativity of the model for the generated inference (choose from Obvious / Somewhat-Creative / Out-of-the-box) i.e. provided the image and inference type, the participant indicates whether the generated inference was apparent from the content of the image (obvious) or the model could generate something beyond what the image conveyed (out-of-the-box).
\end{itemize}

We report Cohen’s kappa statistic as a measurement of inter-annotator agreement \cite{mchugh2012interrater}. The results are reported in Table \ref{tab:user}, and we observe substantial to significant agreement across most inference dimension.

\begin{table}[ht!]

\begin{tabular}{@{}cll@{}}
\toprule
\textbf{Inference (Prompt)} & \multicolumn{1}{c}{\textbf{\begin{tabular}[c]{@{}c@{}}Correctness\\ Kappa\end{tabular}}} & \multicolumn{1}{c}{\textbf{\begin{tabular}[c]{@{}c@{}}Creativity\\ Kappa\end{tabular}}} \\
\midrule 
Pre-condition (Pp2) & 0.78 & 0.76 \\
Effect (Pe4) & 0.68 & 0.66 \\
Goals (Pg4) & 0.74 & 0.83 \\
Before event (Pb3) & 0.81 & 0.81 \\
After event (Pa3) & 0.71 & 0.77 \\ \bottomrule
\end{tabular}
\caption{Human evaluation results: Kappa scores based on the model's ability to generate correct and creative inference about actions along various dimensions}
\label{tab:user}
\end{table}

\section{Conclusion}

In gist, an ability to reason about actions is crucial for Artificial Intelligent (AI) agents in closing the gap between human-level and machine performance in NLP, vision, and robotics tasks. In this work, we emphasize that beyond complex and multi-step physical interactions, actions have tightly coupled commonsense aspects associated with them, which include pre-conditions, executability, effects, high-level goals, and temporal dependency. We set up a pipeline with multiple existing syntactic and semantic NLP tools to automatically collect commonsense data from cooking videos. We formulate this problem as a vision-language task where provided an image, the system will generate text inferences around aforementioned aspects of actions, inspired by \cite{park2020visualcomet}. From our ablations, we show that integration between visual and textual commonsense reasoning is crucial in achieving the best performance. Our experimental results indicate that there is still a large room for improvement in this research area. We believe this work will encourage new research avenues for improved learning and reasoning about fine-grained commonsense. Systems equipped with such a capability will interact with us in more natural ways, will have a better understanding of the world, and will be better partners for humans in performing day-to-day tasks. 

\section*{Limitations}

In this paper, we attempted to automatically collect a dataset based on existing datasets \cite{ZhXuCoAAAI18, ZhLoCoBMVC18}, which are freely available for the research purposes. Although they provide youtube url from which they collected the data, they do not distribute the actual videos. Therefore, we had to use other open-source python libraries \href{https://github.com/yt-dlp/yt-dlp}{yt-dlp} and \href{https://pypi.org/project/youtube-transcript-api/}{youtube-transcript-api} to collect the desirable data. Among 1790 videos in train+val partition of \cite{ZhXuCoAAAI18}, we could not access 189 videos as they were no longer available on YouTube. Therefore, our dataset collection pipeline heavily relies on availability of such resources and requires some manual effort initially to set up the pipeline. Once developed, our pipeline is easily resuable and can be extended to other domains. Moreover, we used several low-level NLP components \citep{Otmazgin2022FcorefFA, honnibal2020spacy, liu2019roberta} to automate and extract the necessary information. The quality and failures of such tools can heavily impact the quality of our collected data. Therefore we advise thorough manual verification of the collected data using our pipeline. Another limitation of the proposed model is that it cannot capture inferences related to situations that might occur during action execution which are not intended. For example, a person may hurt themselves while cutting an object or an object is overcooked/burnt if kept on the stove for longer than their cook time. Our model cannot generate such unintended inferences related to actions. We acknowledge this limitation and plan to address in future.


\section*{Ethics Statement}
In this paper, we work with cooking videos available over YouTube and process them along with annotations from YouCook2 \cite{ZhXuCoAAAI18} dataset to automatically gather data for commonsense reasoning. Later, we use pre-trained GPT3-base \cite{brown2020language} to generate textual commonsense inferences at test time. For decoding, we use nucleus sampling \cite{holtzman2019curious} with p=0.9, in order to generate diverse text inferences. As our efforts were limited to the cooking domain, there was no content with offensive language or known discrimination concerning racial/gender aspects in our collected data as well as model generations. However, several challenges concerning GPT-3 (including hallucinations and harmful content generation) are well-known and text generation results might be affected depending on the input prompts.

\bibliography{emnlp2023}

\begin{thebibliography}{72}
\expandafter\ifx\csname natexlab\endcsname\relax\def\natexlab#1{#1}\fi

\bibitem[{Anderson et~al.(2018)Anderson, Wu, Teney, Bruce, Johnson,
  S{\"{u}}nderhauf, Reid, Gould, and van~den Hengel}]{mattersim}
Peter Anderson, Qi~Wu, Damien Teney, Jake Bruce, Mark Johnson, Niko
  S{\"{u}}nderhauf, Ian~D. Reid, Stephen Gould, and Anton van~den Hengel. 2018.
\newblock \href
  {http://openaccess.thecvf.com/content\_cvpr\_2018/html/Anderson\_Vision-and-Language\_Navigation\_Interpreting\_CVPR\_2018\_paper.html}
  {Vision-and-language navigation: Interpreting visually-grounded navigation
  instructions in real environments}.
\newblock In \emph{In CVPR}.

\bibitem[{Auer et~al.(2007)Auer, Bizer, Kobilarov, Lehmann, Cyganiak, and
  Ives}]{auer2007dbpedia}
S{\"o}ren Auer, Christian Bizer, Georgi Kobilarov, Jens Lehmann, Richard
  Cyganiak, and Zachary Ives. 2007.
\newblock Dbpedia: A nucleus for a web of open data.
\newblock In \emph{The Semantic Web: 6th International Semantic Web Conference,
  2nd Asian Semantic Web Conference, ISWC 2007+ ASWC 2007, Busan, Korea,
  November 11-15, 2007. Proceedings}, pages 722--735. Springer.

\bibitem[{Baral(2010)}]{baral2010reasoning}
Chitta Baral. 2010.
\newblock Reasoning about actions and change: from single agent actions to
  multi-agent actions.
\newblock In \emph{In KR}.

\bibitem[{Battaglia et~al.(2013)Battaglia, Hamrick, and
  Tenenbaum}]{battaglia2013simulation}
Peter~W Battaglia, Jessica~B Hamrick, and Joshua~B Tenenbaum. 2013.
\newblock Simulation as an engine of physical scene understanding.
\newblock \emph{In NAS}.

\bibitem[{Bian et~al.(2023)Bian, Han, Sun, Lin, Lu, and He}]{bian2023chatgpt}
Ning Bian, Xianpei Han, Le~Sun, Hongyu Lin, Yaojie Lu, and Ben He. 2023.
\newblock \href {http://arxiv.org/abs/2303.16421} {Chatgpt is a knowledgeable
  but inexperienced solver: An investigation of commonsense problem in large
  language models}.

\bibitem[{Bisk et~al.(2018)Bisk, Shih, Choi, and Marcu}]{Bisk2018LearningIS}
Yonatan Bisk, Kevin~J. Shih, Yejin Choi, and Daniel Marcu. 2018.
\newblock Learning interpretable spatial operations in a rich 3d blocks world.
\newblock In \emph{In AAAI}.

\bibitem[{Bisk et~al.(2020)Bisk, Zellers, Gao, Choi et~al.}]{bisk2020piqa}
Yonatan Bisk, Rowan Zellers, Jianfeng Gao, Yejin Choi, et~al. 2020.
\newblock Piqa: Reasoning about physical commonsense in natural language.
\newblock In \emph{In AAAI}.

\bibitem[{Boratko et~al.(2020)Boratko, Li, O’Gorman, Das, Le, and
  McCallum}]{boratko2020protoqa}
Michael Boratko, Xiang Li, Tim O’Gorman, Rajarshi Das, Dan Le, and Andrew
  McCallum. 2020.
\newblock Protoqa: A question answering dataset for prototypical common-sense
  reasoning.
\newblock In \emph{Proceedings of the 2020 Conference on Empirical Methods in
  Natural Language Processing (EMNLP)}, pages 1122--1136.

\bibitem[{Bosselut et~al.(2019)Bosselut, Rashkin, Sap, Malaviya, Celikyilmaz,
  and Choi}]{bosselut2019comet}
Antoine Bosselut, Hannah Rashkin, Maarten Sap, Chaitanya Malaviya, Asli
  Celikyilmaz, and Yejin Choi. 2019.
\newblock Comet: Commonsense transformers for automatic knowledge graph
  construction.
\newblock In \emph{Proceedings of the 57th Annual Meeting of the Association
  for Computational Linguistics}, pages 4762--4779.

\bibitem[{Brown et~al.(2020)Brown, Mann, Ryder, Subbiah, Kaplan, Dhariwal,
  Neelakantan, Shyam, Sastry, Askell et~al.}]{brown2020language}
Tom Brown, Benjamin Mann, Nick Ryder, Melanie Subbiah, Jared~D Kaplan, Prafulla
  Dhariwal, Arvind Neelakantan, Pranav Shyam, Girish Sastry, Amanda Askell,
  et~al. 2020.
\newblock Language models are few-shot learners.
\newblock \emph{Advances in neural information processing systems},
  33:1877--1901.

\bibitem[{Dalvi et~al.(2019)Dalvi, Tandon, Bosselut, Yih, and
  Clark}]{dalvi2019everything}
Bhavana Dalvi, Niket Tandon, Antoine Bosselut, Wen-tau Yih, and Peter Clark.
  2019.
\newblock Everything happens for a reason: Discovering the purpose of actions
  in procedural text.
\newblock In \emph{In EMNLP-IJCNLP}.

\bibitem[{Davis and Marcus(2015)}]{davis2015commonsense}
Ernest Davis and Gary Marcus. 2015.
\newblock Commonsense reasoning and commonsense knowledge in artificial
  intelligence.
\newblock \emph{In Communications of ACM}.

\bibitem[{Deng et~al.(2009)Deng, Dong, Socher, Li, Li, and
  Fei-Fei}]{deng2009imagenet}
Jia Deng, Wei Dong, Richard Socher, Li-Jia Li, Kai Li, and Li~Fei-Fei. 2009.
\newblock Imagenet: A large-scale hierarchical image database.
\newblock In \emph{2009 IEEE conference on computer vision and pattern
  recognition}, pages 248--255. Ieee.

\bibitem[{Denkowski and Lavie(2014)}]{denkowski2014meteor}
Michael Denkowski and Alon Lavie. 2014.
\newblock Meteor universal: Language specific translation evaluation for any
  target language.
\newblock In \emph{Proceedings of the ninth workshop on statistical machine
  translation}, pages 376--380.

\bibitem[{Gao et~al.(2018)Gao, Yang, Chai, and Vanderwende}]{gao2018action}
Qiaozi Gao, Shaohua Yang, Joyce Chai, and Lucy Vanderwende. 2018.
\newblock What action causes this? towards naive physical action-effect
  prediction.
\newblock In \emph{In ACL}.

\bibitem[{Girdhar and Ramanan(2020)}]{girdhar2020cater}
Rohit Girdhar and Deva Ramanan. 2020.
\newblock {CATER: A diagnostic dataset for Compositional Actions and TEmporal
  Reasoning}.
\newblock In \emph{ICLR}.

\bibitem[{Gokhale et~al.(2019)Gokhale, Sampat, Fang, Yang, and
  Baral}]{Gokhale2019CVPRWorkshops}
Tejas Gokhale, Shailaja Sampat, Zhiyuan Fang, Yezhou Yang, and Chitta Baral.
  2019.
\newblock Cooking with blocks : A recipe for visual reasoning on image-pairs.
\newblock In \emph{In CVPR Workshops}.

\bibitem[{He et~al.(2017)He, Gkioxari, Doll{\'a}r, and Girshick}]{he2017mask}
Kaiming He, Georgia Gkioxari, Piotr Doll{\'a}r, and Ross Girshick. 2017.
\newblock Mask r-cnn.
\newblock In \emph{Proceedings of the IEEE international conference on computer
  vision}, pages 2961--2969.

\bibitem[{He et~al.(2016)He, Zhang, Ren, and Sun}]{he2016deep}
Kaiming He, Xiangyu Zhang, Shaoqing Ren, and Jian Sun. 2016.
\newblock Deep residual learning for image recognition.
\newblock In \emph{Proceedings of the IEEE conference on computer vision and
  pattern recognition}, pages 770--778.

\bibitem[{Holtzman et~al.(2019)Holtzman, Buys, Du, Forbes, and
  Choi}]{holtzman2019curious}
Ari Holtzman, Jan Buys, Li~Du, Maxwell Forbes, and Yejin Choi. 2019.
\newblock The curious case of neural text degeneration.
\newblock \emph{arXiv preprint arXiv:1904.09751}.

\bibitem[{Hong et~al.(2021)Hong, Lan, Pang, Guo, and
  Cheng}]{hong2021transformation}
Xin Hong, Yanyan Lan, Liang Pang, Jiafeng Guo, and Xueqi Cheng. 2021.
\newblock Transformation driven visual reasoning.
\newblock In \emph{In CVPR}.

\bibitem[{Honnibal et~al.(2020)Honnibal, Montani, Van~Landeghem, Boyd
  et~al.}]{honnibal2020spacy}
Matthew Honnibal, Ines Montani, Sofie Van~Landeghem, Adriane Boyd, et~al. 2020.
\newblock spacy: Industrial-strength natural language processing in python.

\bibitem[{Huang et~al.(2019)Huang, Le~Bras, Bhagavatula, and
  Choi}]{huang2019cosmos}
Lifu Huang, Ronan Le~Bras, Chandra Bhagavatula, and Yejin Choi. 2019.
\newblock Cosmos qa: Machine reading comprehension with contextual commonsense
  reasoning.
\newblock In \emph{Proceedings of the 2019 Conference on Empirical Methods in
  Natural Language Processing and the 9th International Joint Conference on
  Natural Language Processing (EMNLP-IJCNLP)}, pages 2391--2401.

\bibitem[{Juba(2016)}]{JMLR:v17:13-584}
Brendan Juba. 2016.
\newblock \href {http://jmlr.org/papers/v17/13-584.html} {Integrated common
  sense learning and planning in pomdps}.
\newblock \emph{Journal of Machine Learning Research}, 17(96):1--37.

\bibitem[{Kingma and Ba(2014)}]{kingma2014adam}
Diederik~P Kingma and Jimmy Ba. 2014.
\newblock Adam: A method for stochastic optimization.
\newblock \emph{arXiv preprint arXiv:1412.6980}.

\bibitem[{Li et~al.(2022)Li, Kuncoro, Hoffmann, de~Masson~d{'}Autume, Blunsom,
  and Nematzadeh}]{li-etal-2022-systematic}
Xiang~Lorraine Li, Adhiguna Kuncoro, Jordan Hoffmann, Cyprien
  de~Masson~d{'}Autume, Phil Blunsom, and Aida Nematzadeh. 2022.
\newblock \href {https://aclanthology.org/2022.emnlp-main.812} {A systematic
  investigation of commonsense knowledge in large language models}.
\newblock In \emph{Proceedings of the 2022 Conference on Empirical Methods in
  Natural Language Processing}, pages 11838--11855, Abu Dhabi, United Arab
  Emirates. Association for Computational Linguistics.

\bibitem[{Liu et~al.(2019)Liu, Ott, Goyal, Du, Joshi, Chen, Levy, Lewis,
  Zettlemoyer, and Stoyanov}]{liu2019roberta}
Yinhan Liu, Myle Ott, Naman Goyal, Jingfei Du, Mandar Joshi, Danqi Chen, Omer
  Levy, Mike Lewis, Luke Zettlemoyer, and Veselin Stoyanov. 2019.
\newblock Roberta: A robustly optimized bert pretraining approach.
\newblock \emph{arXiv preprint arXiv:1907.11692}.

\bibitem[{Luo et~al.(2021)Luo, Sampat, Tallman, Zeng, Vancha, Sajja, and
  Baral}]{luo2021just}
Man Luo, Shailaja~Keyur Sampat, Riley Tallman, Yankai Zeng, Manuha Vancha,
  Akarshan Sajja, and Chitta Baral. 2021.
\newblock ‘just because you are right, doesn’t mean i am wrong’:
  Overcoming a bottleneck in development and evaluation of open-ended vqa
  tasks.
\newblock In \emph{Proceedings of the 16th Conference of the European Chapter
  of the Association for Computational Linguistics: Main Volume}, pages
  2766--2771.

\bibitem[{Lv et~al.(2020)Lv, Guo, Xu, Tang, Duan, Gong, Shou, Jiang, Cao, and
  Hu}]{lv2020graph}
Shangwen Lv, Daya Guo, Jingjing Xu, Duyu Tang, Nan Duan, Ming Gong, Linjun
  Shou, Daxin Jiang, Guihong Cao, and Songlin Hu. 2020.
\newblock Graph-based reasoning over heterogeneous external knowledge for
  commonsense question answering.
\newblock In \emph{Proceedings of the AAAI conference on artificial
  intelligence}, volume~34, pages 8449--8456.

\bibitem[{McCarthy(1963)}]{mccarthy1963situations}
John McCarthy. 1963.
\newblock Situations, actions, and causal laws.
\newblock Technical report, STANFORD UNIV CA DEPT OF COMPUTER SCIENCE.

\bibitem[{McCarthy and Hayes(1969)}]{McCarthy1969-MCCSPP}
John McCarthy and Patrick Hayes. 1969.
\newblock Some philosophical problems from the standpoint of artificial
  intelligence.
\newblock In B.~Meltzer and Donald Michie, editors, \emph{Machine Intelligence
  4}, pages 463--502. Edinburgh University Press.

\bibitem[{McCarthy et~al.(1960)}]{mccarthy1960programs}
John McCarthy et~al. 1960.
\newblock \emph{Programs with common sense}.
\newblock RLE and MIT computation center Cambridge, MA, USA.

\bibitem[{McHugh(2012)}]{mchugh2012interrater}
Mary~L McHugh. 2012.
\newblock Interrater reliability: the kappa statistic.
\newblock \emph{Biochemia medica}, 22(3):276--282.

\bibitem[{Miller(1995)}]{miller1995wordnet}
George~A Miller. 1995.
\newblock Wordnet: a lexical database for english.
\newblock \emph{Communications of the ACM}, 38(11):39--41.

\bibitem[{Moore(2013)}]{moore2013development}
Chris Moore. 2013.
\newblock \emph{The development of commonsense psychology}.
\newblock Psychology Press.

\bibitem[{Murugesan et~al.(2021)Murugesan, Atzeni, Kapanipathi, Shukla,
  Kumaravel, Tesauro, Talamadupula, Sachan, and Campbell}]{murugesan2021text}
Keerthiram Murugesan, Mattia Atzeni, Pavan Kapanipathi, Pushkar Shukla, Sadhana
  Kumaravel, Gerald Tesauro, Kartik Talamadupula, Mrinmaya Sachan, and Murray
  Campbell. 2021.
\newblock Text-based rl agents with commonsense knowledge: New challenges,
  environments and baselines.
\newblock In \emph{Proceedings of the AAAI Conference on Artificial
  Intelligence}, volume~35, pages 9018--9027.

\bibitem[{Otmazgin et~al.(2022)Otmazgin, Cattan, and
  Goldberg}]{Otmazgin2022FcorefFA}
Shon Otmazgin, Arie Cattan, and Yoav Goldberg. 2022.
\newblock F-coref: Fast, accurate and easy to use coreference resolution.
\newblock In \emph{AACL}.

\bibitem[{Papineni et~al.(2002)Papineni, Roukos, Ward, and
  Zhu}]{papineni2002bleu}
Kishore Papineni, Salim Roukos, Todd Ward, and Wei-Jing Zhu. 2002.
\newblock Bleu: a method for automatic evaluation of machine translation.
\newblock In \emph{Proceedings of the 40th annual meeting of the Association
  for Computational Linguistics}, pages 311--318.

\bibitem[{Park et~al.(2019)Park, Darrell, and Rohrbach}]{park2019viewpoint}
Dong~Huk Park, Trevor Darrell, and Anna Rohrbach. 2019.
\newblock Robust change captioning.
\newblock In \emph{In ICCV}.

\bibitem[{Park et~al.(2020)Park, Bhagavatula, Mottaghi, Farhadi, and
  Choi}]{park2020visualcomet}
Jae~Sung Park, Chandra Bhagavatula, Roozbeh Mottaghi, Ali Farhadi, and Yejin
  Choi. 2020.
\newblock Visualcomet: Reasoning about the dynamic context of a still image.
\newblock In \emph{Computer Vision--ECCV 2020: 16th European Conference,
  Glasgow, UK, August 23--28, 2020, Proceedings, Part V 16}, pages 508--524.
  Springer.

\bibitem[{Patel et~al.(2022)Patel, Gokhale, Baral, and
  Yang}]{Patel2022CRIPPVQACR}
Maitreya Patel, Tejas Gokhale, Chitta Baral, and Yezhou Yang. 2022.
\newblock Cripp-vqa: Counterfactual reasoning about implicit physical
  properties via video question answering.
\newblock In \emph{Conference on Empirical Methods in Natural Language
  Processing}.

\bibitem[{Rajpurkar et~al.(2018)Rajpurkar, Jia, and Liang}]{rajpurkar2018know}
Pranav Rajpurkar, Robin Jia, and Percy Liang. 2018.
\newblock Know what you don’t know: Unanswerable questions for squad.
\newblock In \emph{Proceedings of the 56th Annual Meeting of the Association
  for Computational Linguistics (Volume 2: Short Papers)}, pages 784--789.

\bibitem[{Rashkin et~al.(2018)Rashkin, Sap, Allaway, Smith, and
  Choi}]{rashkin2018event2mind}
Hannah Rashkin, Maarten Sap, Emily Allaway, Noah~A Smith, and Yejin Choi. 2018.
\newblock Event2mind: Commonsense inference on events, intents, and reactions.
\newblock In \emph{Proceedings of the 56th Annual Meeting of the Association
  for Computational Linguistics (Volume 1: Long Papers)}, pages 463--473.

\bibitem[{Ren et~al.(2015)Ren, He, Girshick, and Sun}]{ren2015faster}
Shaoqing Ren, Kaiming He, Ross Girshick, and Jian Sun. 2015.
\newblock Faster r-cnn: Towards real-time object detection with region proposal
  networks.
\newblock \emph{Advances in neural information processing systems}, 28.

\bibitem[{Sakaguchi et~al.(2021)Sakaguchi, Bras, Bhagavatula, and
  Choi}]{sakaguchi2021winogrande}
Keisuke Sakaguchi, Ronan~Le Bras, Chandra Bhagavatula, and Yejin Choi. 2021.
\newblock Winogrande: An adversarial winograd schema challenge at scale.
\newblock \emph{Communications of the ACM}, 64(9):99--106.

\bibitem[{Sampat et~al.(2021)Sampat, Kumar, Yang, and
  Baral}]{sampat2021clevr_hyp}
Shailaja~Keyur Sampat, Akshay Kumar, Yezhou Yang, and Chitta Baral. 2021.
\newblock Clevr\_hyp: A challenge dataset and baselines for visual question
  answering with hypothetical actions over images.
\newblock In \emph{In NAACL:HLT}.

\bibitem[{Sap et~al.(2019{\natexlab{a}})Sap, Le~Bras, Allaway, Bhagavatula,
  Lourie, Rashkin, Roof, Smith, and Choi}]{sap2019atomic}
Maarten Sap, Ronan Le~Bras, Emily Allaway, Chandra Bhagavatula, Nicholas
  Lourie, Hannah Rashkin, Brendan Roof, Noah~A Smith, and Yejin Choi.
  2019{\natexlab{a}}.
\newblock Atomic: An atlas of machine commonsense for if-then reasoning.
\newblock In \emph{In AAAI}.

\bibitem[{Sap et~al.(2019{\natexlab{b}})Sap, Rashkin, Chen, Bras, and
  Choi}]{Sap2019SocialIC}
Maarten Sap, Hannah Rashkin, Derek Chen, Ronan~Le Bras, and Yejin Choi.
  2019{\natexlab{b}}.
\newblock Social iqa: Commonsense reasoning about social interactions.
\newblock In \emph{Conference on Empirical Methods in Natural Language
  Processing}.

\bibitem[{Schuler(2005)}]{schuler2005verbnet}
Karin~Kipper Schuler. 2005.
\newblock \emph{VerbNet: A broad-coverage, comprehensive verb lexicon}.
\newblock University of Pennsylvania.

\bibitem[{Shridhar et~al.(2020)Shridhar, Thomason, Gordon, Bisk, Han, Mottaghi,
  Zettlemoyer, and Fox}]{shridhar2020alfred}
Mohit Shridhar, Jesse Thomason, Daniel Gordon, Yonatan Bisk, Winson Han,
  Roozbeh Mottaghi, Luke Zettlemoyer, and Dieter Fox. 2020.
\newblock Alfred: A benchmark for interpreting grounded instructions for
  everyday tasks.
\newblock In \emph{In CVPR}.

\bibitem[{Speer et~al.(2017)Speer, Chin, and Havasi}]{speer2017conceptnet}
Robyn Speer, Joshua Chin, and Catherine Havasi. 2017.
\newblock Conceptnet 5.5: An open multilingual graph of general knowledge.
\newblock In \emph{Proceedings of the AAAI conference on artificial
  intelligence}, volume~31.

\bibitem[{Tandon et~al.(2014)Tandon, Melo, and Weikum}]{tandon2014acquiring}
Niket Tandon, Gerard Melo, and Gerhard Weikum. 2014.
\newblock Acquiring comparative commonsense knowledge from the web.
\newblock In \emph{Proceedings of the AAAI Conference on Artificial
  Intelligence}, volume~28.

\bibitem[{Valmeekam et~al.()Valmeekam, Olmo, Sreedharan, and
  Kambhampati}]{valmeekam2022large}
Karthik Valmeekam, Alberto Olmo, Sarath Sreedharan, and Subbarao Kambhampati.
\newblock Large language models still can't plan (a benchmark for llms on
  planning and reasoning about change).
\newblock In \emph{NeurIPS 2022 Foundation Models for Decision Making
  Workshop}.

\bibitem[{Vedantam et~al.(2015)Vedantam, Lawrence~Zitnick, and
  Parikh}]{vedantam2015cider}
Ramakrishna Vedantam, C~Lawrence~Zitnick, and Devi Parikh. 2015.
\newblock Cider: Consensus-based image description evaluation.
\newblock In \emph{Proceedings of the IEEE conference on computer vision and
  pattern recognition}, pages 4566--4575.

\bibitem[{Wang et~al.(2017{\natexlab{a}})Wang, Wu, Shen, Dick, and Van
  Den~Henge}]{wang2017explicit}
Peng Wang, Qi~Wu, Chunhua Shen, Anthony Dick, and Anton Van Den~Henge.
  2017{\natexlab{a}}.
\newblock Explicit knowledge-based reasoning for visual question answering.
\newblock In \emph{Proceedings of the 26th International Joint Conference on
  Artificial Intelligence}, pages 1290--1296.

\bibitem[{Wang et~al.(2017{\natexlab{b}})Wang, Wu, Shen, Dick, and Van
  Den~Hengel}]{wang2017fvqa}
Peng Wang, Qi~Wu, Chunhua Shen, Anthony Dick, and Anton Van Den~Hengel.
  2017{\natexlab{b}}.
\newblock Fvqa: Fact-based visual question answering.
\newblock \emph{IEEE transactions on pattern analysis and machine
  intelligence}, 40(10):2413--2427.

\bibitem[{Wang et~al.(2019)Wang, Wang, Chen, and
  Jin}]{wang-etal-2019-youmakeup}
Weiying Wang, Yongcheng Wang, Shizhe Chen, and Qin Jin. 2019.
\newblock \href {https://doi.org/10.18653/v1/D19-1517} {{Y}ou{M}akeup: A
  large-scale domain-specific multimodal dataset for fine-grained semantic
  comprehension}.
\newblock In \emph{Proceedings of the 2019 Conference on Empirical Methods in
  Natural Language Processing and the 9th International Joint Conference on
  Natural Language Processing (EMNLP-IJCNLP)}, pages 5133--5143, Hong Kong,
  China. Association for Computational Linguistics.

\bibitem[{Wu et~al.(2022)Wu, Lu, Sabharwal, and Mottaghi}]{wu2022multi}
Jialin Wu, Jiasen Lu, Ashish Sabharwal, and Roozbeh Mottaghi. 2022.
\newblock Multi-modal answer validation for knowledge-based vqa.
\newblock In \emph{Proceedings of the AAAI Conference on Artificial
  Intelligence}, volume~36, pages 2712--2721.

\bibitem[{Yagcioglu et~al.(2018)Yagcioglu, Erdem, Erdem, and
  Ikizler-Cinbis}]{yagcioglu2018recipeqa}
Semih Yagcioglu, Aykut Erdem, Erkut Erdem, and Nazli Ikizler-Cinbis. 2018.
\newblock Recipeqa: A challenge dataset for multimodal comprehension of cooking
  recipes.
\newblock In \emph{In EMNLP}.

\bibitem[{Yang et~al.(2021)Yang, Panagopoulou, Lyu, Zhang, Yatskar, and
  Callison-Burch}]{yang2021visual}
Yue Yang, Artemis Panagopoulou, Qing Lyu, Li~Zhang, Mark Yatskar, and Chris
  Callison-Burch. 2021.
\newblock Visual goal-step inference using wikihow.
\newblock In \emph{In EMNLP}.

\bibitem[{Ye et~al.(2023)Ye, Xie, Chen, Xu, Yuan, Zhu, and Liao}]{Ye_2023_CVPR}
Shuquan Ye, Yujia Xie, Dongdong Chen, Yichong Xu, Lu~Yuan, Chenguang Zhu, and
  Jing Liao. 2023.
\newblock Improving commonsense in vision-language models via knowledge graph
  riddles.
\newblock In \emph{Proceedings of the IEEE/CVF Conference on Computer Vision
  and Pattern Recognition (CVPR)}, pages 2634--2645.

\bibitem[{Yeo et~al.(2018)Yeo, Lee, Wang, Choi, Cho, Amplayo, and
  Hwang}]{yeo2018visual}
Jinyoung Yeo, Gyeongbok Lee, Gengyu Wang, Seungtaek Choi, Hyunsouk Cho,
  Reinald~Kim Amplayo, and Seung-won Hwang. 2018.
\newblock Visual choice of plausible alternatives: An evaluation of image-based
  commonsense causal reasoning.
\newblock In \emph{In LREC}.

\bibitem[{Yi et~al.(2020)Yi, Gan, Li, Kohli, Wu, Torralba, and
  Tenenbaum}]{yiclevrer}
Kexin Yi, Chuang Gan, Yunzhu Li, Pushmeet Kohli, Jiajun Wu, Antonio Torralba,
  and Joshua~B Tenenbaum. 2020.
\newblock Clevrer: Collision events for video representation and reasoning.
\newblock In \emph{International Conference on Learning Representations}.

\bibitem[{Yin et~al.(2024)Yin, Wang, Horio, Kawahara, and
  Sekine}]{yin2024should}
Ziqi Yin, Hao Wang, Kaito Horio, Daisuke Kawahara, and Satoshi Sekine. 2024.
\newblock Should we respect llms? a cross-lingual study on the influence of
  prompt politeness on llm performance.
\newblock \emph{arXiv preprint arXiv:2402.14531}.

\bibitem[{Zellers et~al.(2019)Zellers, Bisk, Farhadi, and
  Choi}]{zellers2019recognition}
Rowan Zellers, Yonatan Bisk, Ali Farhadi, and Yejin Choi. 2019.
\newblock From recognition to cognition: Visual commonsense reasoning.
\newblock In \emph{Proceedings of the IEEE/CVF conference on computer vision
  and pattern recognition}, pages 6720--6731.

\bibitem[{Zellers et~al.(2018)Zellers, Bisk, Schwartz, and
  Choi}]{zellers2018swag}
Rowan Zellers, Yonatan Bisk, Roy Schwartz, and Yejin Choi. 2018.
\newblock Swag: A large-scale adversarial dataset for grounded commonsense
  inference.
\newblock In \emph{Proceedings of the 2018 Conference on Empirical Methods in
  Natural Language Processing}, pages 93--104.

\bibitem[{Zhan et~al.(2022)Zhan, Huang, Dong, Cao, and
  Liang}]{zhan2022pathreasoner}
Xunlin Zhan, Yinya Huang, Xiao Dong, Qingxing Cao, and Xiaodan Liang. 2022.
\newblock Pathreasoner: Explainable reasoning paths for commonsense question
  answering.
\newblock \emph{Knowledge-Based Systems}, 235:107612.

\bibitem[{Zhang et~al.(2018)Zhang, Liu, Liu, Gao, Duh, and
  Durme}]{abs-1810-12885}
Sheng Zhang, Xiaodong Liu, Jingjing Liu, Jianfeng Gao, Kevin Duh, and
  Benjamin~Van Durme. 2018.
\newblock Record: Bridging the gap between human and machine commonsense
  reading comprehension.
\newblock \emph{CoRR}, abs/1810.12885.

\bibitem[{Zhong et~al.(2019)Zhong, Tang, Duan, Zhou, Wang, and
  Yin}]{zhong2019improving}
Wanjun Zhong, Duyu Tang, Nan Duan, Ming Zhou, Jiahai Wang, and Jian Yin. 2019.
\newblock Improving question answering by commonsense-based pre-training.
\newblock In \emph{Natural Language Processing and Chinese Computing: 8th CCF
  International Conference, NLPCC 2019, Dunhuang, China, October 9--14, 2019,
  Proceedings, Part I 8}, pages 16--28. Springer.

\bibitem[{Zhou et~al.(2019)Zhou, Khashabi, Ning, and Roth}]{zhou2019going}
Ben Zhou, Daniel Khashabi, Qiang Ning, and Dan Roth. 2019.
\newblock “going on a vacation” takes longer than “going for a walk”: A
  study of temporal commonsense understanding.
\newblock In \emph{Proceedings of the 2019 Conference on Empirical Methods in
  Natural Language Processing and the 9th International Joint Conference on
  Natural Language Processing (EMNLP-IJCNLP)}, pages 3363--3369.

\bibitem[{Zhou et~al.(2018{\natexlab{a}})Zhou, Louis, and Corso}]{ZhLoCoBMVC18}
Luowei Zhou, Nathan Louis, and Jason~J Corso. 2018{\natexlab{a}}.
\newblock \href {http://bmvc2018.org/contents/papers/0070.pdf}
  {Weakly-supervised video object grounding from text by loss weighting and
  object interaction}.
\newblock In \emph{British Machine Vision Conference}.

\bibitem[{Zhou et~al.(2018{\natexlab{b}})Zhou, Xu, and Corso}]{ZhXuCoAAAI18}
Luowei Zhou, Chenliang Xu, and Jason~J Corso. 2018{\natexlab{b}}.
\newblock \href
  {https://www.aaai.org/ocs/index.php/AAAI/AAAI18/paper/view/17344} {Towards
  automatic learning of procedures from web instructional videos}.
\newblock In \emph{AAAI Conference on Artificial Intelligence}, pages
  7590--7598.

\end{thebibliography}
\bibliographystyle{acl_natbib}

\appendix

\end{document}